\pdfoutput=1

\documentclass[11pt]{article}

\usepackage[]{EMNLP2022}

\usepackage{times}
\usepackage{latexsym}
\usepackage{enumitem}
\usepackage{booktabs}
\usepackage[T1]{fontenc}

\usepackage[utf8]{inputenc}

\usepackage{microtype}
\usepackage{inconsolata}
\usepackage{multirow}
\usepackage{graphicx}
\usepackage{amsmath}
\usepackage{amsfonts}
\let\oldhat\hat
    
\renewcommand{\hat}[1]{\oldhat{\mathbf{#1}}}

\newcommand{\ie}{\emph{i.e., }}
\newcommand{\eg}{\emph{e.g., }}

\newcommand{\etc}{\emph{etc}}

\title{Sequential Topic Selection Model with Latent Variable for Topic-Grounded Dialogue}

\author{Xiaofei Wen$^{1,2}$, Wei Wei$^{1,2}$\thanks{\hspace{0.15cm}Corresponding author} \and Xian-Ling Mao$^3$\\
   $^1$Cognitive Computing and Intelligent Information Processing Laboratory, School of Computer \\ Science and Technology, Huazhong University of Science and Technology \\
   $^2$Joint Laboratory of HUST and Pingan Property \& Casualty Research (HPL)\\
   $^3$Department of Computer Science and Technology, Beijing Institute
   of Technology \\
  \texttt{xfwen@hust.edu.cn, weiw@hust.edu.cn, maoxl@bit.edu.cn} \\
}
\begin{document}
\maketitle
\begin{abstract}
 Recently, topic-grounded dialogue system has attracted significant attention due to its effectiveness in predicting the next topic to yield better responses via the historical context and given topic sequence. However, almost all existing topic prediction solutions focus on only the current conversation and corresponding topic sequence to predict the next conversation topic, without exploiting other topic-guided conversations which may contain relevant topic-transitions to current conversation. To address the problem, in this paper we propose a novel approach, named \textbf{S}equential \textbf{G}lobal \textbf{T}opic \textbf{A}ttention (SGTA) to exploit topic transition over all conversations in a subtle way for better modeling post-to-response topic-transition and guiding the response generation to the current conversation. Specifically, we introduce a latent space modeled as a Multivariate Skew-Normal distribution with hybrid kernel functions to flexibly integrate the \textit{global}-level information with sequence-level information, and predict the topic based on the distribution sampling results. We also leverage a topic-aware \textit{prior-posterior} approach for secondary selection of predicted topics, which is utilized to optimize the response generation task. Extensive experiments demonstrate that our model outperforms competitive baselines on prediction and generation tasks.
\end{abstract}

\section{Introduction}
Dialog systems have been widely used in a variety of applications. Recent efforts in dialogue systems aim at improving the diversity of agent responses \citep{zhang2018generating,zhou-etal-2021-learning,ijcai2022p453} and endowing agents with the ability to exploit knowledge, express empathy and retain personality \citep{adiwardana2020towards,roller-etal-2021-recipes,DBLP:journals/tois/WeiLMGZZHF21,DBLP:journals/corr/abs-2208-10816}. However, in many real-world scenarios (\eg conversational recommendation, shopping guide and psychological counseling), conversational agents need to proactively steer the conversation by smoothly transforming the conversation topic into a specified one. Therefore, topic-grounded controllable conversation has recently attracted extensive attention. 

Indeed, topic-grounded controllable dialogue is a task to obtain informative responses through a series of predicted topics and the given context. Recently, topic-grounded conversation works mainly focus on modeling the post-to-response \textbf{topic transition} for predicting the next topic to guide the response generation. Many deep learning based approaches are proposed for the task, which utilize topic similarity information as well as additional knowledge information to model dynamic topic transitions \citep{tang-etal-2019-target,wu-etal-2019-proactive,qin2020dynamic,zhou-etal-2020-towards,zhong2021keyword}. These approaches have achieved encouraging results, but they still face the issues as follows. \textit{First}, some of these methods infer the next turn topic only with current turn topic embedding \citep{tang-etal-2019-target,qin2020dynamic,zhong2021keyword}. However, a topic may be inherently associated with several previous topics. Thus they may suffer from the inability of sufficiently modeling the sequential topic-transitions into the topic embeddings as ignoring the inherent topic dependencies over historical topic sequence. \textit{Second}, almost all previous approaches model the topic transferring information only over the current topic sequence while neglecting the useful topic-transition patterns from other sequences, since frequently co-occurring topics over all topic-sequences are more likely to related to each other. Therefore, fully exploring such information is conceptually advantageous to accurately modeling the topic transition for better infer the next topic.
 


\begin{figure}[t]
  \centering
  \includegraphics[width=1.0\linewidth]{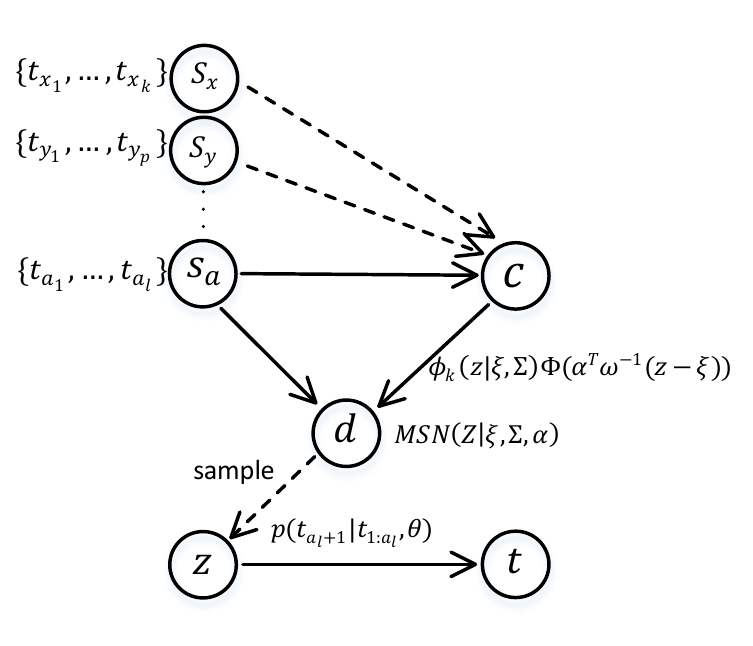}
  \caption{Graphical illustration of topic prediction. Given the current topic sequence $s_a$ and other topic sequences (\eg $s_x$ and $s_y$), the global co-occurrence matrix $c$ is obtained with all topic sequences. Parameters of MSN distribution $d$ are constructed from $s_a$ and $c$ in the latent space, and the next turn topic $t_{a_l+1}$ will be predicted based on the latent variable $z$ obtained after sampling.} 
  \label{fig:graph}
\end{figure}

In this paper, we propose a Transformer-based sequential modeling approach, named \textbf{S}equential \textbf{G}lobal \textbf{T}opic \textbf{A}ttention (SGTA), to fully exploit topic-co-occurrences over all topic sequences for better modeling the post-to-response topic transitions and accurately predicting the next topic to the current conversation. Specifically, SGTA consists of three key elements: topic sequence $s$, global co-occurrence matrix $c$ and latent variable $z$. The relationships among these elements are elaborated with the graphical model in Figure~\ref{fig:graph}. Specifically, we propose a new latent space based on the value variation of the Transformer and use it to model the contextual relation within current topic sequence. The latent space is modeled as a Multivariate Skew-Normal (MSN) distribution \citep{azzalini1996multivariate} due to the flexibility of its \textbf{covariance} parameter to integrate multiple information and the specificity of its \textbf{shape} parameter to control the skewness of the distribution. The covariance parameter $\Sigma$ is designed via task-specific kernel functions for measuring the similarity of pair-wise sequences based on the topic representations. The shape parameter $\alpha$ is used to represent the unique relative relationships of different topics in a sequence. These two parameters are constructed efficiently using current sequence information and other sequences' \textit{global} information. By sampling the MSN distribution on Transformers, we provide a reparameterization of the MSN distribution to enable amortized inference over the latent space. Based on the sampling results, the model simulates the topic transition relationships in the sequence for predicting the next topic in current conversation.

To properly use predicted words and address fluency and controllability issues in generating responses, we leverage a response-based posterior topic distribution to instruct the model to generate responses based on the predicted topics. The \textit{prior} and \textit{posterior} ideas are widely used in response generation tasks \citep{ijcai2019-706}, which is widely-adapted for avoiding to learn the same conversation patterns to each utterance. Concretely, we utilize the \textit{KL-Div Loss} between the \textit{prior} and \textit{posterior} distributions as part of the overall loss, which allows the model to better adapt to response generation by converging the transition scope of the semantic space.

We summarize the contributions of this work as:

\begin{itemize}[leftmargin=*]
	\item To the best of our knowledge, this is the first work that exploits \textit{global}-level topic transitions of all sequence information to learn contextual information for topic-related dialogue.
	\item We propose a unified model to improve the topic prediction performance of current sequence by constructing task-specific MSN distribution using sequential information and global information.
	\item We propose a method that subtly exploits predicted words for response generation and achieves the state-of-the-art on the benchmark dataset compared to competitive baseline methods
\end{itemize}

\section{Related Work}
\begin{figure*}[th]
  \centering
  \includegraphics[width=450px]{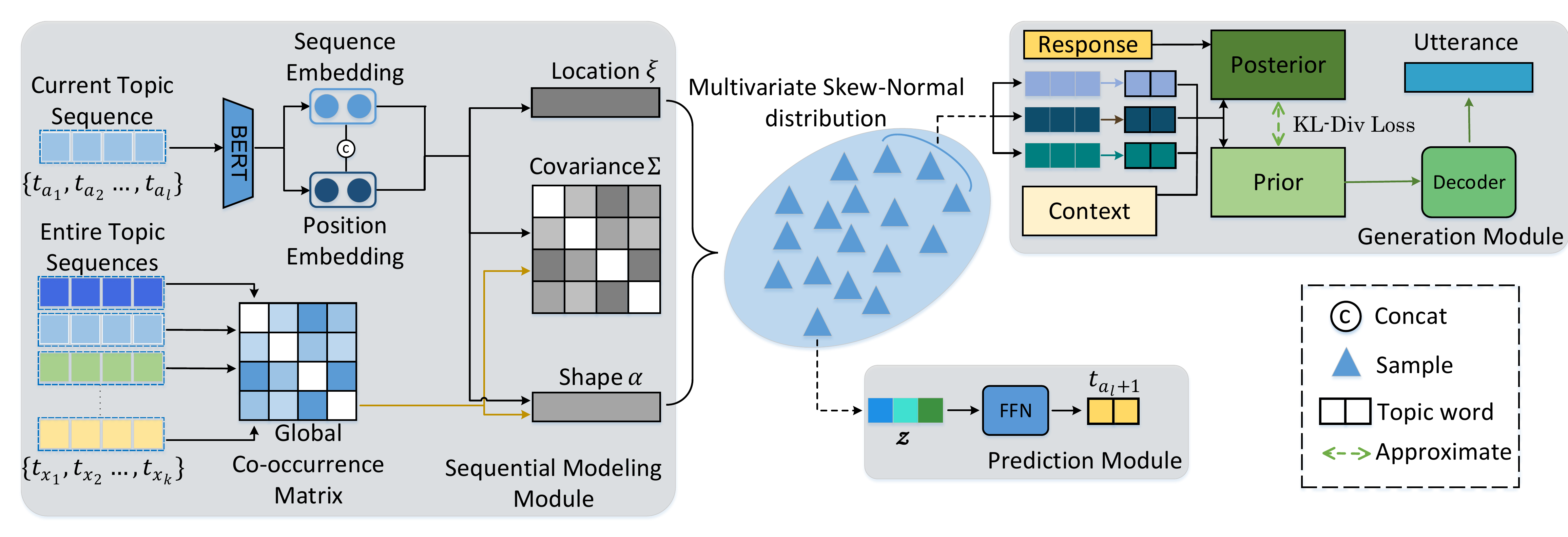}
  \caption{The overall structure of the proposed model SGTA, which consists of (1) a sequential modeling module, (2) a prediction module and (3) a generation module.} 
  \label{fig:SGTA}
\end{figure*}
\noindent\textbf{Topic Transition Conversation} Recently, several studies attempted to build agents that can actively guide conversations by introducing designated target keyword. \citet{tang-etal-2019-target} proposed a next-turn keyword predictor and a rule-based keyword selection strategy to solve the topic transition problem, given the conversation history and the target keyword. \citet{qin2020dynamic} improved \citep{tang-etal-2019-target} by exploiting the semantic knowledge relation among candidate keywords to achieve smooth keyword transition. Very closely, \citet{zhong2021keyword} leverages commonsense knowledge for keyword transition prediction through GNN-based models. Inspired by this category of work for controlled topic conversations, \citet{liu-etal-2020-towards-conversational} first introduce goals in conversational recommendation to enhance the controllability of the conversation process, and \citet{zhou-etal-2020-towards} utilizes Pre-trained Language Model(PrLM) to capture the given current topic sequence to guide conversational recommendation. Our work follows the task definition in (\citealp{zhou-etal-2020-towards}), mainly focusing on topic transition and next-turn topic prediction problem in multi-turn dialogues, given the topic sequence. However, all previous approaches only model the topic-transition information on the current sequence. In contrast, our work learns topic transition information across all sequences to enhance the topic transfer modeling of the current sequence.


\noindent\textbf{Topic-Aware Response Generation} Most of early studies on topic-related dialogue fall into two categories, \ie implicit topic-based \citep{shang-etal-2015-neural,serban2016building,tian-etal-2017-make} and explicit topic-based \citep{xing2017topic,wang2018chat,dziri-etal-2019-augmenting,liu2022incorporating}. The former aims to model contexts at multiple semantic levels (\ie topic, style, \etc) to capture the dynamic conversation topic flow while ignoring the informativeness of responses. The latter generates appropriate utterances on the conversation utterances and given topic information, which relies heavily on manually predefined topic sequences. Due to the excellent performance of the PrLMs on the dialog generation task, \citet{liu-etal-2021-durecdial} and \citet{zhou-etal-2020-towards} choose to encode topic information and context by the PrLM to obtain informative and fluent responses. However, it may suffer from topic-noise problem (\ie raw topic fitting and similar topic selection) when generating responses. In contrast, our work leverages several topics to a \textit{prior} distribution, using \textit{posterior} information on already known topic to guide the generation of \textit{prior} distribution that affects the generation process.


\section{Methodology}
We propose a novel model for Topic-guided Dialogue. Figure~\ref{fig:SGTA} presents the architecture of our model, which comprises three main components: 1) \textit{Global}  Sequential Topic Attention layer; 2) Point-Wise Feed-Forward Network Prediction layer; 3) \textit{Prior} and \textit{Posterior} Response Generation layer. We next present the three components and parameter modeling in detail. 
\subsection{Problem Statement}
Let $\mathcal{C}=\left\{u_{1}, u_{2}, \ldots, u_{|\mathcal{C}|}\right\}$ be a multi-turn conversation; let $\mathcal{S} =\left\{t_{1}, t_{2}, \ldots, t_{l}\right\}$ be the topic sequence of conversation $\mathcal{C}$, and let $\mathcal{T} = \left\{t_{1}, t_{2}, \ldots, t_{|\mathcal{T}|}\right\}$ be all of topics.

Given the conversation $\mathcal{C}$ and topic sequence $\mathcal{S}$, the task of topic guide conversation aims to predict the top-$N$ topics $(1 \leq N \leq |\mathcal{T}| )$ from $\mathcal{T}$ at turn $k$ and generate appropriate response $u_{|\mathcal{C}|+1}$.

\subsection{\textit{Global} Sequential Topic Attention layer}
In this section, we will describe how to perform serialized embedding of topics in Section~\ref{sec-embedding} and how to construct MSN distribution based on sequential information and global information in Section~\ref{sec-MSN}.
\subsubsection{Embedding layer}
\label{sec-embedding}
To obtain a better semantic representation of the topic words in the sequence, we encode the topic sequences using BERT \citep{devlin-etal-2019-bert}; also, to bring the model to aware previous topics information while addressing the sequence, we add position embedding for each topic. We use the latest $n$ topics in $\mathcal{S}$, where $n \geq l$, and we pad constant zero vector to make each sequence length equal. Specifically, the topic embedding matrix is defined as $\mathbf{E} \in \mathbb{R}^{|\mathcal{T}| \times d}$, where $d$ denotes the topic embeddings' dimensionality and $\mathbf{E}$ is the estimated by BERT. We extract the input matrix $\widehat{\mathbf{E}} \in \mathbb{R}^{n \times d}$, where $\widehat{\mathbf{E}}_s = \mathbf{E}_{T_s}$. Inspired by \citet{kang2018self}, we inject a learnable position embedding $\mathbf{P} \in \mathbb{R}^{n \times d}$ into the input embedding:
\begin{equation}
  \widehat{X} = [e_i \parallel p_{l-i+1}]
  \end{equation}
where $e_i \in \widehat{\mathbf{E}}_s$ and $p_i \in \mathbf{P}$. We make the reverse location embedding injection due to the unfixed length of the topic sequence. Compared with the regular forward location, the distance between each topic and the predicted topic contains certain effective information \citep{wang2020global,DBLP:conf/sigir/Zou0MWQ0C22}.

\subsubsection{Multivariate Skew-Normal Distribution}
\label{sec-MSN}
The core of our model is constructed based on Transformer's Multi-Head Attention, we choose Transformer due to its excellent ability in sequence modeling problem \citep{kang2018self,yu2019multi,ji2020sequential}. For the Scaled-Dot Attention in Multi-Head Attention, given the Equation~(\ref{eq:baseatt}):
\begin{equation}
    \text{ Attn }(Q, K, V)= \text{softmax} (\frac{Q K^{T}}{\sqrt{d_{k}}}) V 
  \label{eq:baseatt}
  \end{equation}

Since the key to attention mechanism is querying via the alignment score, we turn the alignment score (\ie scaled dot product of $Q$ and $K^T$) into a latent variable $z$, which allows us to adapt the model to apply the mandatory information flexibly, including \textit{global}-level transition and topic relativities. Despite the superior results demonstrated by multivariate normal distribution in modeling the covariance matrix \citep{fisher2011improved}, which exhibits the intra-sequence topic correlation, due to its forced symmetric shape of the density curve (\ie suffering from modeling skewness), we devise $z$ to follow the Multivariate Skewed Normal (MSN) distribution \citep{azzalini1996multivariate}. According to \citet{azzalini1999statistical}, $z \in \mathbb{R}^{k} $ is continuous with density function as:
\begin{equation}
  f(z)=2 \phi_{k}(z-\xi; \Sigma) \Phi\{\alpha^{T} \omega^{-1}(z-\xi)\}
  \label{MSN-base}
  \end{equation}
where $\Sigma = \omega\psi\omega$ is the covariance matrix and  $\psi \in \mathbb{R}^{k \times k} $ denotes correlation parameter, as well as $\xi = (\xi_{1},...,\xi_{k})^{T}$, $\omega = \text{diag}(\omega_1,...,\omega_k) $ and $\alpha \in \mathbb{R}^{k}$ denote the location, scale and shape parameters respectively. Moreover, $\phi_{k}$ is the $k$-dimensional multivariate normal density with the mean $\xi$ and the covariance $\Sigma$, and $\Phi(\cdot)$ is the $N(0,1)$ distribution function. For clarity, we denote the above distribution obediently as Equation~(\ref{eq:MSN-z}) and follow the original notation:
\begin{equation}
  z \sim MSN(Z \mid \xi, \Sigma, \alpha) 
  \label{eq:MSN-z}  
  \end{equation}

As we need to sample the $\text{softmax}$ parameter values from MSN distribution by which to revise the alignment score, we adapt Equation~(\ref{eq:baseatt}) to derive the following equation:
\begin{equation}
  \begin{array}{r}
  \operatorname{H}=\operatorname{SGTA}(T_{seq}, G_b)=\operatorname{softmax}(Z) V
  \end{array}
  \label{Atten-apply}
  \end{equation}
where $T_{seq}$ denotes the input topic sequence, $G_b$ denotes the \textit{global}-level topic transition co-occurrence matrix, which is constructed by statistically calculating the topic co-occurrence in the whole dataset. Furthermore, $\operatorname{H} = \{h_1,...,h_n\}$ is denoted as hidden layer output with $h_i \in \mathbb{R}^d $. All parameters modeling will be explained in detail in subsequent sections.

\noindent\textbf{Location} The Location $\xi$ represents the mean of the distribution. Considering that we demand the deterministic alignment scores with maximum likelihood to facilitate sampling the MSN distribution, we manipulate the alignment score as:

\begin{equation}
  \xi = \text{LeakyReLU}(\frac{(T_{seq} \mathbf{W}^{Q}_{l})(T_{seq} \mathbf{W}^{K}_{l})}{\sqrt{d}}) 
\end{equation}
To keep the standard deviation positive, we choose $\text{LeakyReLU}$ as the activation function. 

\noindent\textbf{Covariance} The covariance $\Sigma$ indicates the relation between two different topics, which is formed by two subparameters $\omega$ and $\psi$ as we mentioned before. For scale $\omega$, given a topic sequence $\mathcal{S}$ with topic position $\mathcal{S}_p^i$ and $\mathcal{S}_p^j$, we normalize them to $x_i$ and $x_j$ and infer the variance $\omega_i^2, \omega_j^2 \in \mathbb{R}_{+} $ of $z$ respectively, by amortization inference as in Equation~(\ref{eq:leaky}).
\begin{equation}
  \omega_m = \text{LeakyReLU}(\frac{(x_n \mathbf{W}_{\omega}^Q)(x_m \mathbf{W}_{\omega}^K)}{\sqrt{d}})
  \label{eq:leaky}
\end{equation}

As for correlation $\psi$, we adopt the kernel function approach for the mixing calculation, since it can efficiently and nonlinearly compute the inner product of samples in the feature space and calculate high-dimensional distance measures. It's well known that proximal distance metric illustrates proximate relationship. Especially, we comprise a hybrid kernel function for building task-aware metric. Different from \citet{tang-etal-2019-target} and \citet{ji2020sequential}, we consider 
global co-occurrence level topic transition information to construct kernel functions for metrics, nevertheless they build only based on the embedding similarity, which is difficult to obtain sufficient information. We then describe the specific implementation of kernel function.
\begin{description}[leftmargin=*]
  \item[$\bullet$ Co-occurrence kernel] is determined by the number of co-occurrence between pairwise topics, which is linearly dependent with the topic pairs. The co-occurrence kernel is designed as:
\begin{equation}
  k_{co}(x_i,x_j) = \omega_i\omega_j\log(\frac{P_{ij}^{\beta}}{p_ip_j})
\end{equation}
where $P_{ij}$ is the co-occurrence value of $x_i$ and $x_j$ in the common sequence and $p_i$, $p_j$ are their individual occurrence values. $\beta$ is the factor that determines the impact degree of the co-occurrence value $P_{ij}$.
  \item[$\bullet$ Topic pair kernel] is determined by topic transition pattern pair representation, which is strongly correlated with the topic pairs association. The topic pair kernel $k_{tp}(x_i,x_j)$ is designed as:
  \begin{equation}
    k_{tp} =\omega_i\omega_j (\exp(-\gamma{\| x_i - x_j \| }^2)
    + x_i x_j)
  \end{equation}
which combines both Gaussian kernel and . In particular, the Gaussian kernel is primarily used to characterize the similarity between samples, and $\gamma > 0$ is the unique hyperparameter of Gaussian kernel function.
\end{description}

To make full use of the above two part information, we design the above two kernel functions to be summed as shown in Equation~(\ref{final_kernel}):
\begin{equation}
  k(x_i,x_j) = k_{co}(x_i,x_j) + \eta rk_{tp}(x_i,x_j)
  \label{final_kernel}
\end{equation}
where $r = \text{softmax}(x\mathbf{W}_x + \mathbf{b}_x)$, $\mathbf{W}_x \in \mathbb{R} ^{d \times d}$, $\mathbf{b}_x \in \mathbb{R}^d$ and $\eta$ denote the learnable parameters. After relation modeling, we set the correlation matrix $\psi_{ij} = \frac{k(x_i,x_j)}{\omega_i\omega_j}$ and substitute $\psi$ into $\Sigma=\omega\psi\omega$ to infer the \textit{covariance} $\Sigma$. 

\noindent\textbf{Shape} The shape parameter $\alpha$ reflects the relation of each topic in the sequence with the last topic. It contains the consideration for relative position information, while this correlation helps the model to learn the implicit transition pattern between \textit{topic}-level and \textit{position}-level relations. Specifically, we let $\alpha_i = s_{n-i}\frac{\widehat{\alpha}_i }{\text{max}(\widehat{\alpha})}$ represents correlation between the final topic $t_n$ and topic $t_i$, where $\widehat{\alpha}$ is a ratio parameter which mirrors the estimation and consideration at \textit{topic}-level, while $s_{n-i}$ is a relative scaling parameter with \textit{position}-level information.

We use the co-occurrence matrix $G_b$ to calculate the ratio parameter $\widehat{\alpha}_i$, divided into intra-sequence level and global level parts, representing the sum of linear arrangement from $t_i$ to $t_n$ and influence factors summation of the top-$m$ frequently co-occurring topics with $t_i$, respectively. $g_{i,j}$ is the $i$-th row and $j$-th column value of $G_b$ (\ie the value of $t_i$ and $t_j$). The detailed formula for $\widehat{\alpha}_i$ is given below:
\begin{equation}
  \widehat{\alpha}_i = \underbrace{\sum_{j = 1}^{n} g_{i,j}g_{j,n}}_{\textit{intra-sequence level}} + \underbrace{\sum_{l=1}^{m} g_{i,k_l}g_{k_l,n}}_{\textit{global level}}
  \label{shape}
\end{equation}

Equation~(\ref{shape}) calculates $\widehat{\alpha}_i$ by utilizing the sum of two dot products, which exhibits the correlation between $t_i$ and $t_n$, where $g_{i,k_l}$ stands for the $l$-th element after decreasing the order of all $g_{i,k}$ values with $k \in |\mathcal{T}|$, and $m$ is an adjustable variable. Inspired by \citet{ji2020sequential}, we utilize the average of the pairwise dot product sums of the remaining topics in the sequence to replace $g_{j,j}$ since it is an invalid value in matrix $G_b$, which can be expressed as:
\begin{equation}
  g_{j,j} \Longleftarrow \underbrace{\frac{\sum_{p \in \{1,...,n\}\setminus{j}} g_{j,p} }{n-1}}_{\textit{average of remaining co-occurrence}}
\end{equation}

Similar to $\omega$ in section \textit{Covariance}, we define the scaling parameter $s_{n-i}$ as Eq.~(\ref{eq:leaky_s}):
\begin{equation}
  s_{n-i} = \text{LeakyReLU}(\frac{(x_n \mathbf{W}_s^Q)(x_i \mathbf{W}_s^K)|n-i|}{\sqrt{d}})
  \label{eq:leaky_s}
\end{equation}
\noindent\textbf{Loss function}
Given the above construction of the distribution parameters, the latent variable $z$ needs to be inferred according to MSN distribution. Following \citet{kingma2013auto}, $z$ can be inferred by optimizing the lower bound on the evidence of Jensen's inequality for the marginal logarithm $p(y_n)$ while predicting the $(n + 1)$-th topic. We present the loss function as shown in Eq.~(\ref{ELBO}).
\begin{equation}
  \begin{aligned}
    \mathcal{L}_{z}(\theta) &=\mathbb{E}_{z}\left[\log p\left(y_{n} \mid z\right)\right] \\
  & \leq \log \int p\left(y_{n} \mid z\right) p(z) d z=\log p\left(y_{n}\right)
  \end{aligned}
  \label{ELBO}
  \end{equation}
We also design the co-occurrence loss $L_{rank}$ by listwise ranking loss \citep{cao2007learning} to match topic-co-occurrence relevance and ranking consistency. The total loss $\mathcal{L}$ is defined as the sum of the co-occurrence loss $\mathcal{L}_{rank}$ and the latent variable loss $\mathcal{L}_{z}$: 
\begin{equation}
    \mathcal{L} = \mathcal{L}_z + \delta\mathcal{L}_{rank}
\end{equation}
where $\delta$ is a hyperparameter. We also use the reparameterization trick \citep{NIPS2015_bc731692} to ensure that the model is trainable.

\subsection{Point-Wise Feed-Forward Network Prediction layer}
We apply the PointWise Feed-Forward network in Transformer to the output of the model and incorporate location dependent information. The pointwise feedforward network consists of two linear layers and the activation layer. The output $F=\{\text{FFN}(h_1),...,\text{FFN}(h_d)\}$. We also stacks $b$ self-attentive blocks to adaptively and hierarchically extract previous consumed topic information and learn complex topic transition patterns.

After the above adapted query attention, we predict the next topic (given the first $n$ topics) based on $F_n^{(b)}$:
\begin{equation}
  r_{i,n} = F_n^{(b)}\mathbf{E}_{i}
\end{equation}
where $r_{i,n}$ is the relevance of topic $t_i$ being the next topic given the first $n$ topics (\ie $t_1,...,t_n$), and $\mathbf{E}_{i}$ is the BERT embedding of topic $t_i$. Empirically, high relevance represents a compact transfer relation, so we use ranking $r$ to make predictions for next topic $t_i$.

\subsection{\textit{Prior} and \textit{Posterior} Response Generation layer}

After obtaining the predicted topic words, it is essential to employ them flexibly to generate smooth response for completing exhaustive dialogue interaction. Most of the previous methods select topics for generating responses based on the similarity between the previous topic and next turn's topic or context, which can be regarded as the topic \textit{prior} distribution, however there are actual conversation scenarios in which multiple candidate topics are pertinent for the previous topic transitions (\eg \textit{
movie}$\rightarrow$\textit{music} and \textit{movie}$\rightarrow $\textit{friend}). Both \textit{music} and \textit{friend} can be considered as the next topic word for \textit{movie}. Nevertheless, with the \textit{posterior} distribution constructed from query and response, model can be more empirical in selecting the appropriate topics for generation.

Particularly, during training, we design that the posterior distribution $p(t=t_i|q,r)$ containing the response information $r$ is approximated by the prior distribution $p(t=t_i|q)$ containing only historical information $q$, which includes context $c$ and history topic sequence $t_{1:n-1}$ (\ie $q = \{c,t_{1:n-1}\}$). We evaluate the approximation loss with \textit{KL-Div Loss}, as shown in the following equation:
\begin{equation}
  \mathcal{L}_{KL}(\theta) = \sum_{i = 1}^{N}p(t=t_i|q,r) \log \frac{p(t=t_i|q,r)}{p(t=t_i|q)}
  \label{KLD}
\end{equation}

The posterior distribution can guide the model to generate natural responses according to secondary selected topic,and converging the semantic transition range while generation. Compared to the previous method, $BOWLoss$ serves as a part of the overall loss for evaluating each word of the predicted responses, while allowing for a better fit to the true responses, and meanwhile ensure the controllability of the generated utterances. Following \citet{ijcai2019-706}, the $BOWLoss$ is defined as:
\begin{equation}
  \mathcal{L}_{BOW}(\theta) = -\mathbf{E}_{t_i \sim p(t|c,r)} \sum_{j=1}^{m} \log p(r_j|t_i)
  \label{BOW}
\end{equation}
where $\theta$ denotes the model parameters in Eq.~(\ref{KLD}) \& Eq.~(\ref{BOW}). Additionally, we apply Transformer \citep{vaswani2017attention} as the decoder of our model to generate outstanding and convincing responses. For name consistency, we denote this generation method via SGTA in the whole paper.
\section{Experiment}
\begin{table*}[ht]
  \centering
  \scalebox{1.0}{
\begin{tabular}{l|ccc|ccc}
   \toprule
   \multicolumn{1}{l|}{\multirow{2}*{\textbf{Model}}}      & \multicolumn{3}{c|}{\textit{with target}} & \multicolumn{3}{c}{\textit{without target}} \\
   \cmidrule{2-7}
   \multicolumn{1}{c|}{} & \textbf{Hit@1}       & \textbf{Hit@3}       & \textbf{Hit@5}     & \textbf{Hit@1}       & \textbf{Hit@3}      & \textbf{Hit@5}     \\
   \midrule
   PMI                 & $0.0349$    & $0.0927$    & $0.1290$    &-       & -      & -      \\
   Kernel  &$0.0418$ & $0.0957$ & $0.1125$&-&-&- \\
   DKRN &$0.4015$ &$0.4821$ &$0.5068$ &-&-&- \\
   CKC &$0.5914$ &$0.7857$&$0.8265$&$0.3269$&$0.4505$&$0.4961$ \\
   \midrule
   MGCG                      & $0.5861$       & $0.7528$       & $0.8094$       & $0.3157$       & $0.4386$      & $0.4483$      \\
   Conversation-BERT                      & $0.6072$   & $0.7893$    & $0.8105$    & $0.2952$    & $0.4099$   & $0.4585$  \\
   Topic-BERT                      & $0.6104$ & $0.7966$ & $0.8147$  & $0.4185$   & $0.5520$  & $0.5971$  \\
   TG-CRS &$0.6128$ &$0.8157$ &$0.8302$ & $0.4352$ &$0.5726$ &$0.6291$ \\
   \midrule
   SGTA         & $\mathbf{0.6208^*}$       & $\mathbf{0.8523^*}$      & $\mathbf{0.8671^*}$       & $\mathbf{0.4487^*}$      & $\mathbf{0.6030^*}$      & $\mathbf{0.6546^*}$      \\
   \bottomrule
\end{tabular}}
\caption{Automatic evaluation of topic predictions task. \textbf{Bold} text indicates the best result. Significant improvements compared to the best baseline are marked with $*$ (t-test, $p$ < 0.05).}
\label{table_Hit_result}
\end{table*}
\subsection{Dataset}
\noindent\textbf{TG-ReDial dataset} The TG-ReDial dataset \citep{zhou-etal-2020-towards} is composed of 10,000 conversations between seekers and recommenders. It contains a total of 129,392 utterances from 1,482 users, covering 2,571 topics. The dataset is constructed in a topic-guided manner, where each conversation includes a topic sequence as well as a conversation target, and both parties communicate sequentially according to the topic sequence to accomplish the target and thus achieve the recommendation. It is notable that the conversation targets are \textit{\textbf{optional}} in our task setting as additional information in the model input, and we will give a detailed experimental comparison later. On average, each conversation in the TG-ReDial dataset has 7.9 topics and a utterance contains 19 words.

\subsection{Implementation details}
We implemented SGTA and related baseline experiments in PyTorch. The default parameters for all experiments are set as follows: we set the batch size to 16 and the embedding size is set to 768. We used two self-attention blocks and one head for sequence modeling following \citet{kang2018self}. It is worth noting that we constructed the co-occurrence matrix only from the training set data. We used Adam optimizer \citep{kingma2015adam} with a learning rate initialized to $1e^{-5}$ and the dropout rate is set to 0.1.

\subsection{Baselines}
Our baselines for assess the performance of topic prediction and response generation come in two groups:

\textit{Topic Guide Prediction:} (1) PMI and Kernel \citep{tang-etal-2019-target} employ the forcing strategy of selecting keywords with higher similarity to the target. (2) DKRN \citep{qin2020dynamic} introduces a graph routing mechanism for keyword search based on \citet{tang-etal-2019-target}. (3) MGCG \citep{liu-etal-2020-towards-conversational} adopt a completion judgment mechanism in the exploration of topic sequence. (4) CKC \citep{zhong2021keyword} follows the work of (1) and (2), introduces the commonsense graph and utilizes the GNN method to select keywords. (5) Conversation-BERT, Topic-BERT and TG-CRS \citep{zhou-etal-2020-towards} respectively input the context, historical topic sequence and the concatenation of them to BERT for encoding and predicting. 

The above baselines are briefly divided into two sub-categories, \textit{similarity}-based (\ie PMI, Kernel, DKRN and CKC) and \textit{sequence}-based (\ie MGCG, Conversation-BERT, Topic-BERT and TG-CRS). The former predicts based on the similarity of next turn topic and target while the latter predicts based on the topic sequence and context.

\begin{table*}[t]
  \centering
  \scalebox{0.95}{
\begin{tabular}{l|ccccc|cc}
   \toprule
   \multicolumn{1}{l|}{\multirow{2}*{\textbf{Model}}}      & \multicolumn{5}{c|}{\textit{Automatic}} & \multicolumn{2}{c}{\textit{Human}} \\
   \cmidrule{2-8}
   \multicolumn{1}{c|}{} & \textbf{PPL}    & \textbf{BLEU-1}   & \textbf{BLEU-2}       & \textbf{Distinct-1}     & \textbf{Distinct-2 }       & \textbf{Fluency}      & \textbf{Informativeness}     \\
   \midrule
   KBRD  &$28.022$&$0.221$ & $0.028$ & $0.004$&$0.008$&$1.16$&$1.30$ \\
   KGSF &$40.758$&$0.239$ &$0.042$ &$0.015$ &$0.064$&$1.35$&$1.35$ \\
   MGCG  &$12.386$& $0.256$  & $0.061$& $0.012$&$0.041$& $1.31$ & $1.26$ \\
   Transformer           &    $32.856$       & $0.287$       & $0.071$       & $0.013$       & $0.083$       & $1.35$      & $1.31$      \\
   TG-CRS &$\mathbf{7.223}$&$0.280$ &$0.065$ &$0.021$ & $0.094$ &$\mathbf{1.42}$ &$1.43$ \\
   \midrule
   SGTA         &$8.539$& $\mathbf{0.301^*}$       & $\mathbf{0.080^*}$      & $\mathbf{0.023^*}$       & $\mathbf{0.124^*}$      & $1.41$      & $\mathbf{1.47}$      \\
   \bottomrule
\end{tabular}}
\caption{Automatic evaluation and human evaluation of response generation task. \textbf{Bold} text indicates the best result. Significant improvements compared to the best baseline are marked with $*$ (t-test, $p$ < 0.05).}
\label{table_Generation_result}
\end{table*}

\textit{Topic Response Generation:} (1) KBRD \citep{chen2019towards} enhances the transformers' decoder with keywords searched on the knowledge graph. (2) KGSF \citep{zhou2020improving} incorporates both word-oriented and entity-oriented knowledge graphs to enhance response. (3) MGCG adopts multi-task learning framework to predict topics and generates responses with posterior distributions. (4) Transformer \citep{vaswani2017attention} is a widely used multi-head attention encoder-decoder structure. (5) TG-CRS utilizes the predicted word concating context input to GPT-2 \citep{radfordlanguage} to generate responses.

\subsection{Evaluation metrics}
\noindent\textbf{Automatic evaluation} To evaluate the topic prediction task, following the previous work \citep{tang-etal-2019-target,zhou-etal-2020-towards,liu-etal-2020-towards-conversational}, we adopt Hit@$k(k=1,3,5)$ as metric for ranking all the possible topics (topic keywords recall at position $k$ in all keywords). To evaluate the response generation task, following \citet{zhou-etal-2020-towards}, we adopt perplexity (PPL), BLEU-$n(n=1,2)$ and Distinct-$n(n=1,2)$ for examining the fluency and informativeness of the responses.

\noindent\textbf{Human evaluation} For generation tasks, it is necessary to employ manual evaluation to test the ability of the models to make topic-related informative responses. We randomly select 100 dialogues from our model and baseline, recruit four annotators to evaluate several models in two aspects, \ie \textit{fluency} and \textit{informativeness}. The former measures whether the generated responses are fluent, while the latter measures whether the system introduces relevant topics in response. Score ranges from 0 to 2, and the final average score is calculated as the result.

\subsection{Performance on Topic Prediction}
Table~\ref{table_Hit_result} shows the experimental results of topic prediction task. PMI and Kernel do not perform well, since they cannot consider the topic sequence context. CKC performs significantly better than other \textit{similarity}-based baselines as it exploits the external information in the commonsense knowledge graphs. We notice that TG-CRS outperforms the other baselines since it jointly models dialogue context, topic sequence, and user profile. SGTA gives an increase of 1.3\%, 4.5\%, and 4.4\% over TG-CRS in terms of Hit@1/3/5 respectively with consideration of target information, as SGTA effectively leverages the topic-transition patterns in other topic sequences. It should be clarified that since user-level information is not taken into account in our task, our experimental set-up that contains the target \textit{differs from} the original set-up, where we concatenate the target word embedding as additional information with the topic sequence.
It is easy to figure that conversation topic transfer tends to be more uncontrollable without considering target information, where only CKC can make predictions in the \textit{similarity}-based approach due to its having sufficient external information as a prediction base. Moreover, in the \textit{sequence}-based methods, Topic-BERT is fairly better than Conversation-BERT, which shows the effectiveness of topic sequence. SGTA benefits from better topic sequence modeling, improves over the best baseline by 3.1\%, 5.4\% and 4.2\% on Hit@1/3/5, respectively. 


\subsection{Performance on Response Generation}

Table~\ref{table_Generation_result} exhibits the generation performance of automatic and manual metrics on the TG-REDIAL dataset. For automatic evaluation, we find that TG-CRS performs best in the baselines, indicating the robustness of the pre-trained model for the generation task. KGSF outperforms KBRD and MGCG in terms of diversity because it combines KG-enhanced item and word embedding to generate the utterance. Transformer performs better on the BLEU metric due to its word-by-word attention mechanism. SGTA performs best on BLEU and Distinct using a prior and posterior approximation without depending on pre-trained language models, improving 7.5\%, 23\%, 9.5\% and 32\% over TG-CRS at BLEU-1/2 as well as Distinct-1/2 respectively, though PPL is slightly weaker than TG-CRS's effect. As for the manual results, SGTA improves 2.8\% compared to TG-CRS in the informativeness dimension, owing to the fact that the posterior approach makes model learn reasonable topic choices easily, and also SGTA's performance is in the first tier in terms of fluency.

\subsection{Ablation study}
\begin{table}[ht]
  \centering
  \scalebox{0.9}{
\begin{tabular}{l|ccc}
   \toprule
   \multicolumn{1}{l|}{\multirow{2}*{\textbf{Model}}}      & \multicolumn{3}{c}{\textit{with target}} \\
   \cmidrule{2-4}
   \multicolumn{1}{c|}{} & \textbf{Hit@1}       & \textbf{Hit@3}       & \textbf{Hit@5}      \\
   \midrule
   SGTA         & $\mathbf{0.6208^*}$       & $\mathbf{0.8523^*}$      & $\mathbf{0.8671^*}$     \\
   \midrule
   w/o global                & $0.5852$ & $0.7863$ & $0.8121$    \\
   w/o intra-pos            & $0.6081$   & $0.8034$    & $0.8256$ \\
   w/o sequen                     & $0.5749$  & $0.7687$  & $0.7905$ \\
   w/o MSN                            &$0.5895$ &$0.7993$ &$0.8203$ \\
   \bottomrule
\end{tabular}}
\caption{Results of the ablation study for topic prediction task.}
\label{table_Ablation}
\end{table}

\begin{table}[ht]
  \centering
  \scalebox{0.9}{
\begin{tabular}{l|ccc}
   \toprule
   \multicolumn{1}{l|}{\multirow{2}*{\textbf{Model}}}      & \multicolumn{3}{c}{\textit{with target}} \\
   \cmidrule{2-4}
   \multicolumn{1}{c|}{} & \textbf{Hit@1}       & \textbf{Hit@3}       & \textbf{Hit@5}      \\
   \midrule
   SGTA         & $\mathbf{0.6208^*}$       & $\mathbf{0.8523^*}$      & $\mathbf{0.8671^*}$     \\
   \midrule
   w/o MSN                            &$0.5895$ &$0.7993$ &$0.8203$ \\
   w/o MSN-$\xi$                & $0.6097$ & $0.8107$ & $0.8343$    \\
   w/o MSN-$\Sigma$            & $0.5964$   & $0.8039$    & $0.8267$ \\
   w/o MSN-$\alpha$                    & $0.5982$  & $0.8051$  & $0.8255$ \\
   
   \bottomrule
\end{tabular}}
\caption{Results of the ablation study for MSN parameters.}
\label{table_ab_t}
\end{table}
We conduct ablation study to assess the importance of global co-occurrence (\textit{w/o global}), intra-sequence position (\textit{w/o intra-pos}), topic sequence context (\textit{w/o sequen}), as well as the MSN distribution (\textit{w/o MSN}), and results are presented in Table~\ref{table_Ablation}. Concretely, after removing global topic co-occurrence information (\textit{w/o global}), the average reduction in our model effect is about for 6.6\%, which demonstrates the desirability of extracting global information. On this basis, the effect of the model is improved by only removing the intra-sequence position information (\textit{w/o intra-pos}), but it is still lower than our model by 4.2\%. Moreover, the experiment to remove the MSN distribution (\textit{w/o MSN}) show that the information selection with skewness is effective under the premise of reasonable modeling parameters. Finally, experiments with the removing of sequential transformer modeling (\textit{w/o sequen}) show the key role of both in the model structure.

We also conduct ablation study to assess the performance of MSN distribution three parameters: Location $\xi$ (\textit{w/o MSN-$\xi$}), Covariance $\Sigma$ (\textit{w/o MSN-$\Sigma$}) and Shape $\alpha$ (\textit{w/o MSN-$\alpha$}) and Table~\ref{table_ab_t} shows the results. We use the experimental design with the parameter set to zero, and each ablation study represents the effect of removing only the individual parameter separately. Specifically, shape parameter $\alpha$ and the covariance parameter $\Sigma$ perform more important effects in the distribution construction.

\section{Conclusion}
In this paper, we focus on topic-grounded controllable dialogue tasks. We propose a new approach named SGTA. It models the latent space as a MSN distribution utilizing global information, intra-sequence semantic and position information, which allows the model to better integrate the relationships between information and make topic predictions based on the results of distribution sampling. We also utilize a \textit{prior-posterior} distribution approach to generate a new topic response. Extensive experiments on TG-ReDial dataset show that SGTA achieves state-of-the-art performance on prediction and generation tasks.

\section*{Acknowledgement}
This work was supported in part by the National Natural Science Foundation of China under Grant No. 62276110, Grant No.61772076, in part by CCF-AFSG Research Fund under Grant No.RF20210005, and in part by the fund of Joint Laboratory of HUST and Pingan Property \& Casualty Research (HPL). The authors would also like to thank the anonymous reviewers for their comments on improving the quality of this paper. 

\section*{Limitations}
Our work suffers from the following limitations: (1) lack of experimental results on a more topic conversation dataset. Although there are few existing topic conversation datasets and TG-ReDial is an explicit dataset at the topic level, experiments on more datasets are necessary for the generalizability of the method. (2) Our topic prediction has yet to be improved in mining global information at the dataset level. After sampling and verifying the experimental results we believe that the extraction by topic word co-occurrence has some shortcomings, such as words with more co-occurrence may not appear under the current round, which requires the integration of context as well as semantics.


\bibliography{emnlp2022.bbl}
\bibliographystyle{acl_natbib}

\clearpage




\end{document}